# SINGLE UPPER LIMB POSE ESTIMATION METHOD BASED ON IMPROVED STACKED HOURGLASS NETWORK

Gang Peng, Yuezhi Zheng, Jianfeng Li*, Jin Yang, Zhonghua Deng

Key Laboratory of Image Processing and Intelligent Control, Ministry of Education,
School of Artificial Intelligence and Automation, Huazhong University of Science and Technology, Wuhan 430074, China

At present, most high-accuracy single-person pose estimation methods have high computational complexity and insufficient real-time performance due to the complex structure of the network model. However, a single-person pose estimation method with high real-time performance also needs to improve its accuracy due to the simple structure of the network model. It is currently difficult to achieve both high accuracy and real-time performance in single-person pose estimation. For use in human–machine cooperative operations, this paper proposes a single-person upper limb pose estimation method based on an end-to-end approach for accurate and real-time limb pose estimation. Using the stacked hourglass network model, a single-person upper limb skeleton key point detection model was designed. Deconvolution was employed to replace the up-sampling operation of the hourglass module in the original model, solving the problem of rough feature maps. Integral regression was used to calculate the position coordinates of key points of the skeleton, reducing quantization errors and calculations. Experiments showed that the developed single-person upper limb skeleton key point detection model achieves high accuracy and that the pose estimation method based on the end-to-end approach provides high accuracy and real-time performance.

## 1. Introduction

In human–machine cooperative operations, the robot must estimate the pose of a single upper limb of the operator accurately and in real time and provide the pose information for trajectory prediction (Hu et al., 2019) of the upper limb, to enable safe human–machine cooperation without collision (Zlatanski et al., 2019).

When using a convolutional neural network (Koziarski and Cyganek, 2018; Ning et al., 2020) for pose estimation, the position coordinates of the key points of the human skeleton are directly regressed using the input image or video. Toshev and Szegedy (2014) proposed a human pose estimation method based on the AlexNet network framework, which represents the human pose estimation problem as the regression of key points of the human skeleton. Subsequently, Fan et al. (2015) proposed a dual-source deep convolution neural network in which the local parts were combined with the overall view, enabling more accurate human pose estimation. However, the direct regression method is not suitable for low-resolution images, has high computational complexity, and has difficulty ensuring the accuracy of the position coordinates of the key points. Therefore, based on the thermodynamic diagram method (Hu and Ramanan, 2015; Lifshitz et al., 2016; Tompson et al., 2015), Pfister et al. (2015) proposed a deeper convolutional neural network for human pose estimation. This method successfully transforms the problem of human pose estimation into one of human skeleton key point detection. Later, Yang et al. (2016) proposed an end-to-end human pose estimation framework, in which a deep convolution neural network was combined with a tree structure diagram model, but the calculation efficiency of the diagram model was low and the real-time performance required improvement. Further, Wei et al.

---

(2016) proposed a convolutional pose machine model based on a convolutional neural network to remedy the low efficiency of the graph model and to make reasonable use of the spatial position information, texture information, and intermediate constraint relationship of the human body structure. The model abandons the graph method, uses a large convolution kernel to enhance the receptive field, and uses multi-stage regression to improve the accuracy of human pose estimation. However, the large convolution kernel causes high computer resource consumption. To overcome this issue, Newell et al. (2016) proposed a multi-stage regression stacked hourglass network model in which the multi-scale feature method was used to capture the spatial position information of each key point of the human skeleton, yielding the position coordinates of each key point. This method greatly improved the receptive field and reduced the amount of calculation. Subsequently, Chu et al. (2017) designed a novel hourglass residual unit, in which the stacked hourglass network model and attention mechanism were combined to solve the problem of incorrect estimation under a complex background or self-occlusion. Simultaneously, Yang et al. (2017) used the pyramid residual module based on the stacked hourglass network model and studied the multi-branch network weight initialization method to enhance the accuracy of human skeleton key point detection.

To achieve the accuracy and real-time performance requirements of single upper limb pose estimation in human–robot collaboration, this paper proposes a single upper limb pose estimation method based on an end-to-end approach. A single-person upper limb skeleton key point detection model was designed using a stacked hourglass network model with high accuracy and real-time performance, and the hourglass module and human skeleton key point coordinate calculation method in the detection model were improved to increase the detection accuracy. Experiments showed that the improved single upper limb skeleton key point detection model is effective that and the single upper limb pose estimation method based on the end-to-end approach provides high accuracy and real-time performance.

The specific content and structure of this paper are as follows:

Section 1 introduces the background and significance of the study, expounds the current research status of single-person pose estimation, and analyzes the key technical issues involved in this paper.

Section 2 introduces two methods for estimating the pose of a single person upper limb.

Section 3 describes the design and improvement of a single-person upper limb skeleton key point detection model.

Section 4 verifies the improved detection model described in Section 3, and conducts comparative experiments on two methods of single-person upper limb pose estimation, and analyzes the experimental results.

Section 5 summarizes the main work of this paper, and analyzes the limitations and expectations of the study.

## 2. Single upper limb pose estimation

**2.1. Single upper limb pose estimation method based on end-to-end approach.** The single upper limb pose estimation method based on the end-to-end approach uses a single upper limb skeleton key point detection model. The specific process is as follows:

(1) Compare the numbers of rows and columns in the input image to obtain the larger value M, then fill the input image with a square image with M rows and columns and adjust the filled square image to $256 \times 256$ pixels;

(2) Using the image processed in the first step as the input of the single upper limb skeleton key point detection model, detect the single upper limb skeleton key points in the image, and obtain the position coordinates of the key points;

(3) Connect the single upper limb skeleton key points obtained in the second step in a single upper limb pose structure model according to the positions of and connections between the skeleton key points.

The flow chart of the single upper limb pose estimation method based on the end-to-end approach is shown in Fig. 1.

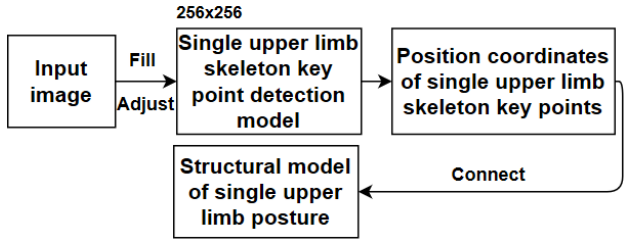

Fig. 1. Flow chart of single-person upper limb pose estimation method based on end-to-end approach.

**2.2. Single upper limb pose estimation method based on cascade method.** The single upper limb pose estimation based on the cascade method uses a human detector and single upper limb skeleton key point detection model. The specific process is as follows:

(1) Use the YOLOv3 network model as a human body detector to detect a single human body in the image and to obtain a human body detection frame; after expanding the human body detection frame by 15%, cut the input image according to the expanded human body detection frame to obtain the body image;

(2) Compare the numbers of rows and columns of the human image acquired in the first step to obtain the larger value M, then fill the human image with a square image with M rows and columns, and adjust the filled square image to $256 \times 256$ pixels;

(3) Using the human body image processed in the second step as the input of a single upper limb skeleton key point detection model, detect the single-person upper limb skeleton key points in the image and obtain the position coordinates of the key points;

(4) Connect the single upper limb skeleton key points obtained in the third step into a single upper limb post-structure model according to the positions of and connections between the skeleton key points.

The flow chart of the single upper limb pose estimation method based on the cascade method is shown in Fig. 2.

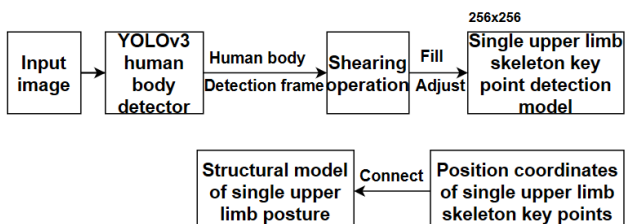

Fig. 2. Flow chart of single upper limb pose estimation method based on cascade method.

## 3. Single upper limb skeleton key point detection

The following introduces the method of designing the single upper limb skeleton key point detection model.

**3.1. Original design of detection model.** The process of designing the single upper limb skeleton key point detection model is as follows.

**3.1.1. Structural design of the model.** This study was based on the model structure of a stacked hourglass network, and a single upper limb skeleton key point detection model was designed that outputs only seven key points of the human upper limb. These points include left and right wrist, left and right elbow, left and right shoulder, and neck key points. To verify that the stacked hourglass network model composed of eight first-order hourglass modules provides better accuracy and real-time detection performance for the skeleton key points, a number of single upper limb skeleton key point detection models were designed with different numbers and orders of hourglass modules and relevant experiments were performed. Table 1 shows the accuracy and real-time detection performances of single upper limb skeleton key point detection models with different hourglass module numbers and orders. The models are named using the form sh_ [number of hourglass modules] [order of hourglass modules], and each detection model was trained on the MPII data set and fine-tuned using the homemade training set. The times in Table 1 are the processing times of the models.

Table 1. Experimental results of multiple single upper limb skeleton key point detection models.

| Model name | Accuracy PCKh@0.5/% | | | Time/ms |
|---|---|---|---|---|
| | Shoulder | Elbow | Wrist | |
| sh21 | 92.9 | 87.5 | 84.4 | 72 |
| sh22 | 93.6 | 88.0 | 84.9 | 109 |
| sh24 | 94.3 | 88.7 | 85.6 | 168 |
| sh41 | 94.0 | 88.4 | 84.9 | 117 |
| sh42 | 94.5 | 89.0 | 85.6 | 185 |
| sh44 | 95.2 | 89.6 | 86.4 | 293 |

| sh81 | 94.7 | 89.1 | 85.8 | 175 |

It can be seen from Table 1 that as the number and order of hourglass modules increase, the accuracy of the single upper limb skeleton key point detection model increases, but the real-time performance requires improvement. The single upper limb skeleton key point detection model composed of eight one-stage hourglass modules provides reasonable accuracy and real-time performance. The structural flow chart of the single upper limb skeleton key point detection model designed in this study is shown in Fig. 3.

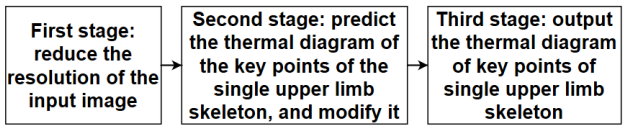

Fig. 3. Structural flow chart of the single upper limb skeleton key point detection model.

It can be seen from Fig. 3 that the structural flow chart of the single upper limb skeleton key point detection model is divided into three stages. The first stage (Fig. 4(a)) is mainly used to convolute the input image, which is then passed through the residual module and lower sampling layer, to reduce the resolution of the input image. The main purpose of the second stage (Fig. 4(b)) is to stack seven modules, including a first-order hourglass module, a residual module, a convolution layer, a batch normalization module, and an activation function layer, to predict the thermal diagram of the key points of the single upper limb skeleton and to revise the thermal diagram continuously. The third stage (Fig. 4(c)) is mainly composed of a first-order hourglass module, a residual module, a convolution layer, a batch normalization module, and an activation function layer. The purpose is to output the thermal diagram of the key points of the single upper limb skeleton after constant correction.

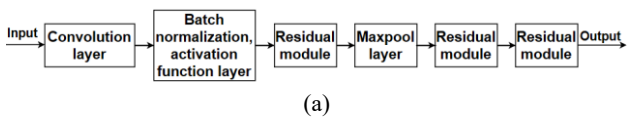
(a)

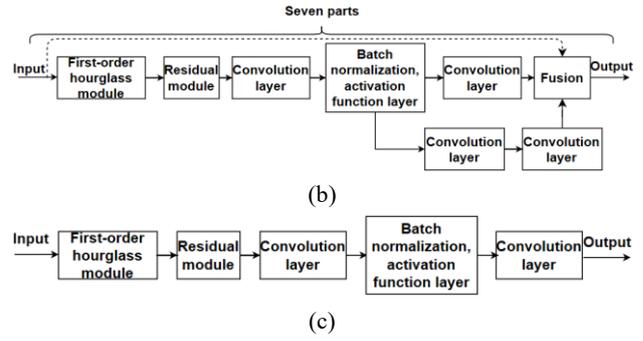
(b)

(c)

Fig. 4. Structure flow charts of the (a) first, (b) second, and (c) third stages of the single upper limb key point detection model.

The structure diagrams of the residual and first-order hourglass modules are shown in Figs. 5(a) and 5(b), respectively.

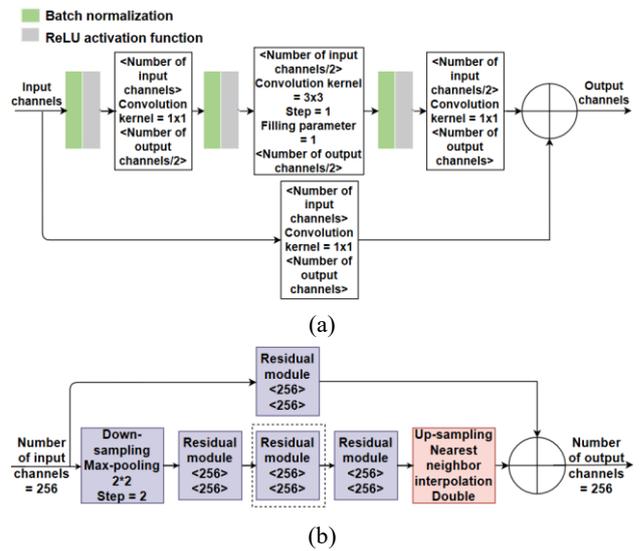

Fig. 5. Structure diagrams of the (a) residual and (b) first-order hourglass modules.

**3.1.2. Loss function.** In the process of training the detection model of the single upper limb bone key points, the mean square error (MSE) is used as the loss function to calculate the error between the predicted and actual thermal diagrams of the upper limb skeletal key points of the human body, so that the hourglass module can be evaluated more accurately:

$$MSE^p = \frac{1}{m}\sum_{n=1}^{m}(\hat{y}_n^p - y_n^p)^2. \quad (1)$$

Here, $m$ is the total number of pixels in the thermal diagram of the single upper limb skeletal key points, $\hat{y}_n^p$ is the probability corresponding to each pixel position $n$ in the predicted thermal diagram of the $p$-th upper limb skeletal key points, and $y_n^p$ is the probability corresponding to each pixel position $n$ in the actual thermal diagram of the $p$-th upper limb skeletal key points.

### 3.1.3. Calculation of position coordinates of key points of bone.
In the single upper limb skeleton key point detection model, the maximum probability method is used to calculate the position coordinates of the skeleton key points. The position corresponding to the pixel with the greatest probability in the thermal diagram of the skeleton key points of the upper limb is taken as the position coordinate of the skeleton key point, which can be expressed as follows:

$$J_k = argmax_p H_k(p). \quad (2)$$

Here, $J_k$ is the position coordinate of the $k$-th skeleton key point of the human upper limb, $p$ is the position in the thermal diagram of the skeleton key point, and $H_k$ is the predicted thermal diagram of the skeleton key point.

With the original design of the single upper limb skeleton key point detection model, the accuracies of the key point detection of the elbow and wrist of the test sample image are 89.1% and 85.8%, respectively. To increase the accuracy, it is necessary to improve the original design of the single upper limb skeleton key point detection model.

### 3.2. Improved detection model design.
The following describes the process of improving the single-person upper limb skeleton key point detection model from two aspects.

### 3.2.1. Improvement of hourglass module.
The hourglass module in the original design of the single upper limb skeleton key point detection model uses the nearest neighbor interpolation method for the up-sampling operation of the feature map. In this method, the content of the feature map is copied directly to expand the feature map, causing the feature map to be rough. Therefore, in this study, we used deconvolution to expand the feature map by slightly improving the up-sampling operation in the hourglass module to obtain a more precise feature map. The specific structure diagram is shown in Fig. 6.

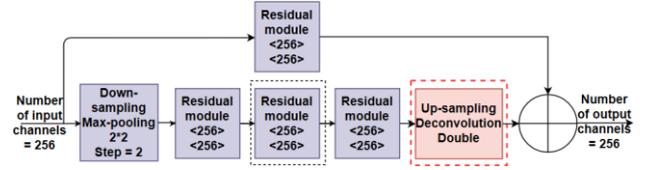

Fig. 6. Structure diagram of improved first-order hourglass module.

The convolution kernel parameter in deconvolution is determined while training the network, and the relation between the dimensions of the input and output characteristic graphs of deconvolution can be expressed as follows:

$$Output_{size} = stride \cdot (Input_{size} - 1) + Kernel_{size} - 2 \cdot padding. \quad (3)$$

$Output_{size}$ is the size of the output feature map, $Input_{size}$ is the size of the input feature map, $Kernel_{size}$ is the size of the convolution kernel, and $padding$ is the fill parameter.

### 3.2.2. Improvement of the skeleton key point coordinate calculation.
Because the original design of the single upper limb skeleton key point detection model uses the maximum probability value to calculate the position coordinates of the skeleton key points, the accuracy of the single upper limb skeleton key point detection model is easily affected by the down-sampling operation. After down-sampling, the resolution of the thermal diagram of the key points of the single upper limb skeleton is much lower than that in the original image, leading to irreversible quantization error. Simultaneously, if the thermal diagram of the key points of the human upper limb skeleton adopts a relatively high resolution, it will cause complex calculations, increased memory consumption, and low real-time performance. Therefore, in this study, integration regression was used instead of the maximum probability method to calculate the position coordinates of the key points of the skeleton, that is, to calculate the integral of all positions in the thermal diagram of the skeleton key points of the upper limb, and the calculation results were taken as the results for the key points of the skeleton, which can be expressed as follows:

$$J_k = \sum_{p_y=1}^{H} \sum_{p_x=1}^{W} p \cdot \frac{e^{H_k(p)}}{\int_{q\in\Omega} e^{H_k(q)}}. \qquad (4)$$

Here, $J_k$ is the position coordinate of the $k$-th skeleton key point of the human upper limb, $p$ is the position of the thermal diagram of the skeleton key point of the human upper limb, and $\Omega$ is the area of the thermal diagram.

In this study, the improved single upper limb skeleton key point detection model was trained on the MPII data set and fine-tuned on the self-made training set. In the self-made test set, the key points of the single upper limb skeleton in the test sample image were detected, and the detection effects are shown in Fig. 7.

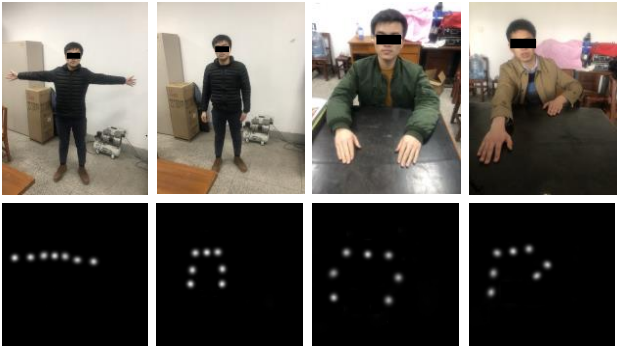

Fig. 7. Detection effects of single upper limb skeleton key point detection model.

## 4. Experimental results and analysis

### 4.1. Setting of experimental parameters and hardware conditions.

**4.1.1. Setting of experimental parameters.** To train and fine-tune the original and improved single upper limb skeleton key point detection models, the training parameters were set as shown in Table 2.

Table 2. Training parameter settings of the single upper limb skeleton key point detection model.

| Parameter name | Set value or method |
|---|---|
| Order of hourglass module | 1 |
| Number of hourglass modules | 8 |
| Optimization method | RMS prop algorithm |
| Initial learning rate | 0.00025 |
| Loss function | MSE |
| Batch size | 8 |
| Epoch parameter | 100 |
| Number of epoch iterations | 1000 |
| Data augmentation method | Random crop, color dither, rotation |

**4.1.2. Hardware conditions in the experiment.** The hardware conditions used in the comparison experiments in this study are shown in Table 3.

Table 3. Hardware conditions in comparison experiments.

| Operating system | Ubuntu16.04 LTS |
|---|---|
| Deep learning framework | Pytorch |
| CPU model | i7-7700K |
| CPU frequency | 4.2GHZ |
| RAM | 32GB |
| Graphics card type | NVIDIA TITAN XP 11G |
| CUDA version | CUDA 8.0 |

### 4.2. Comparison experiments and results analysis.

The comparison experiments performed in this study included single upper limb skeleton key point detection model and single upper limb pose estimation method comparisons and analysis of the experimental results.

**4.2.1. Single upper limb skeleton key point detection model comparison.** To confirm the validity of the improved model, the original model, improved model, cascaded pyramid network model, and convolution attitude machine network model were compared in the same environment with the same test set. Among them, the cascaded pyramid network model and convolution attitude network model were retrained on the MPII dataset and self-made training set. In the contrast experiment, the test sample image was zoomed to the appropriate size and then directly input into the above four detection models. The accuracy curves of four key point detection models of the single upper limb skeleton used to detect the key points of the elbow and wrist in the test sample image are shown in Fig. 8.

· 6 ·

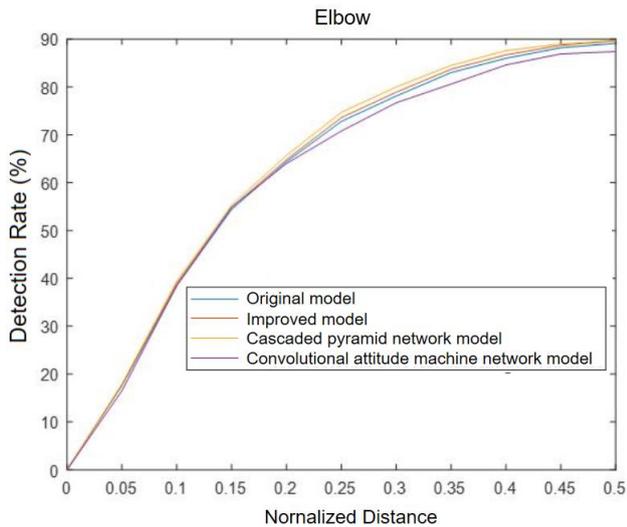

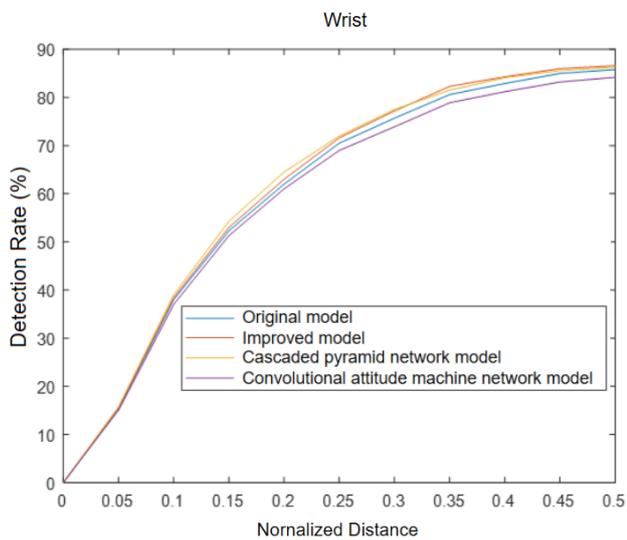

Fig. 8. Accuracy curves of four single upper limb skeleton key point detection models for the (a) elbow and (b) wrist.

Using 0.5 as the threshold value of PCKh, the accuracies and real-time performances of four single upper limb skeleton key point detection models were obtained using test samples and are presented in Table 4. The times in the table are the model processing times.

As shown in Fig. 8 and Table 4, the proposed single-person upper limb skeleton key point detection model improves the detection accuracy. Compared with the cascaded pyramid network model, the key point detection accuracy of the improved model is only slightly different, but the real-time performance is superior. Compared with the convolutional attitude machine network model, the improved model has slightly higher detection accuracy and real-time performance. Therefore, the key point detection model based on the improved stack hourglass network is effective.

Table 4. Experimental results of four single upper limb skeleton key point detection models.

| Model name | Accuracy PCKh@0.5/% | | | Time/ms |
|---|---|---|---|---|
| | Shoulder | Elbow | Wrist | |
| Convolutional attitude machine network | 94.0 | 87.4 | 84.2 | 398 |
| Cascaded pyramid network | 95.6 | 89.8 | 86.3 | 278 |
| Original detection model | 94.7 | 89.1 | 85.8 | 175 |
| Improved detection model | 95.4 | 89.6 | 86.6 | 179 |

**4.2.2. Single upper limb pose estimation method comparison.** In this study, the improved single upper limb skeleton key point detection model was applied in combination with two single upper limb pose estimation methods. For the end-to-end single upper limb pose estimation method, the experiment was conducted according to the steps described in Section 2.1. For the single upper limb pose estimation method in cascade mode, the YOLOv3 network model was fine-tuned on the self-made training set, and the single upper limb pose estimation experiment was performed according to the steps described in Section 2.2. For more convincing comparisons, the original input image was zoomed to the appropriate size and then directly input into the original and improved designs of the single upper limb skeleton key point detection model.

The accuracy curves corresponding to the four methods of detecting the key points of the elbow and wrist in the test sample images are shown in Fig. 9.

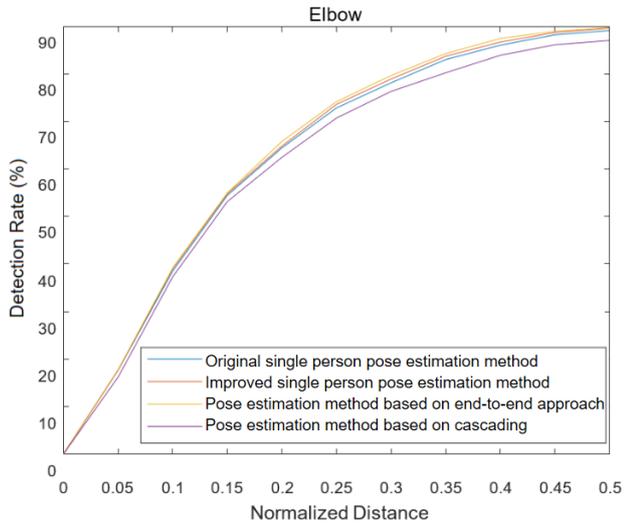

(a)

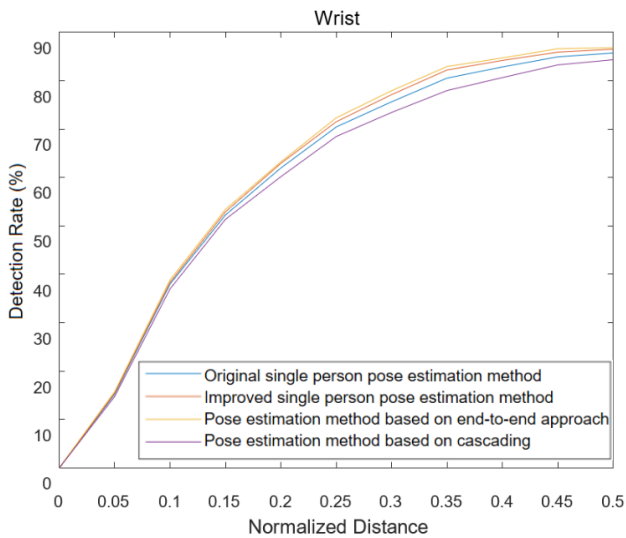

(b)

Fig. 9. Detection accuracy curves of four single upper limb pose estimation methods for the (a) elbow and (b) wrist.

Using 0.5 as the threshold value of PCKh, the accuracies and real-time performances of four single upper limb pose estimation methods for single upper limb skeleton key points were obtained by using test samples and are presented in Table 5.

Table 5. Experimental results of four single upper limb pose estimation methods.

| Method name | Accuracy PCKh@0.5/% | | | Time /ms |
|---|---|---|---|---|
| | Shoulder | Elbow | Wrist | |
| Original pose estimation method | 94.7 | 89.1 | 85.8 | 175 |
| Improved pose estimation method | 95.4 | 89.6 | 86.6 | 179 |
| Pose estimation method based on end-to-end approach | 95.7 | 89.8 | 86.9 | 171 |
| Pose estimation method based on cascading | 93.2 | 87.0 | 84.4 | 214 |

As can be seen from Fig. 9 and Table 5, the end-to-end single upper limb pose estimation method is slightly more accurate than directly scaling the original input image to the appropriate size and inputting it into the detection model. Because the original input image is filled and then adjusted, the original horizontal-to-vertical ratio of the image can be maintained when adjusting the image, so the adjusted image will not be deformed and the accuracy will be slightly better. Compared with the end-to-end single upper limb pose estimation method, the method obtained by cascading a human detector and an improved single upper limb skeleton key point detection model yielded poor accuracy and real-time performance. These poor results were obtained because the human detector based on the YOLOv3 network model could not accurately detect the human body in the original input image and the input image of the single upper limb skeleton key point detection model was based on the human detection results. Therefore, the human body detector incorrectly detected the human body in the original input image and used the wrong human body image as the input for the single upper limb skeleton key point detection model, reducing the detection accuracy. The human body detector also needs time to detect the human body in the original input image.

In general, these results demonstrate that the single upper limb pose estimation method based on the end-to-end approach has better accuracy and real-time performance than the cascade method.

**4.2.3. Experimental results of single-person upper limb pose estimation method.** The experimental results obtained using the single upper limb pose estimation methods based on the end-to-end approach and cascade method are shown in Figs. 10 and 11, respectively; and the results of the pose estimation experiment for each video frame are shown in Fig. 12. It can be seen that the proposed single-person upper limb pose estimation method is feasible and effective.

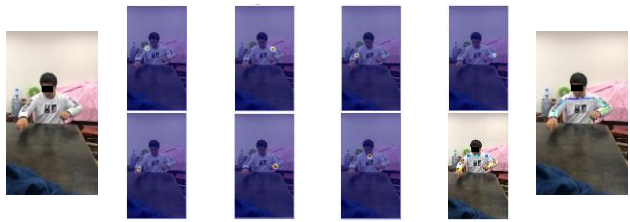

Fig. 10. Experimental results of single upper limb pose estimation based on end-to-end approach.

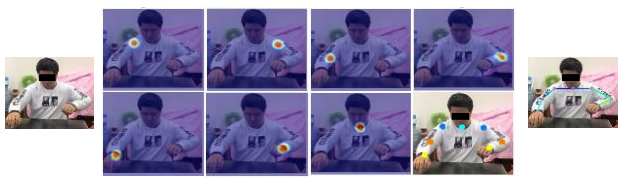

Fig. 11. Experimental results of single upper limb pose estimation based on cascade method.

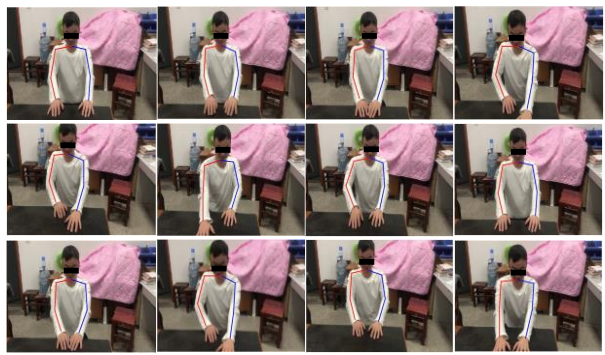

Fig. 12. Effect of single upper limb pose estimation on each frame in the same video.

## 5. Conclusion

Based on previous research on human–machine cooperative operation, an end-to-end single-person upper limb pose estimation method was developed in this study. Using the stacked hourglass network model, a single-person upper limb skeleton key point detection model was designed and the hourglass module and human skeleton key point coordinate calculation method were improved. Experiments confirmed the effectiveness of the improved single upper limb skeleton key point detection model. Compared with single upper limb pose estimation based on the cascade method, the proposed end-to-end single-person upper limb pose estimation method achieves higher accuracy and real-time performance.

This study also has some limitations, which need to be further improved. This study focuses on the estimation method of human upper limb pose, which can provide the pose data for the subsequent trajectory prediction of upper limb movement. This will be the focus of future work.

**Gang Peng** is an associate professor at Huazhong University of Science and Technology and the head of the Intelligent Robots Lab. He received his Ph.D. in engineering from Huazhong University of Science and Technology in 2002 and is engaged in research on industrial robots. Email: penggang@hust.edu.cn.

**Yuezhi Zheng**, Co-First Author, received a master's degree from Huazhong University of Science and Technology, China, in 2019. His research interests include robotics and computer vision.

**Jianfeng Li**, Corresponding Author, received a bachelor's degree in engineering from Nanchang University, China, in 2019. He is currently pursuing his MS degree in control science and engineering at Huazhong University of Science and Technology. His research interests include robotics and computer vision. Email: lijianfeng1218@163.com.

**Jin Yang** received a bachelor's degree in engineering from Tianjin Polytechnic University, China, 2019. He is currently pursuing a master's degree in control science and engineering at Huazhong University of Science and Technology. His research interests include robotic arm control and image processing.

**Zhonghua Deng** graduated from Huazhong University of Science and Technology with a degree in Control Theory and Control Engineering in 1994, and obtained a doctorate in engineering. He is currently a professor in the School of Artificial Intelligence and Automation of Huazhong University of Science and Technology.